\documentclass[sn-basic,iicol,Numbered]{sn-jnl}

\usepackage{graphicx}%
\usepackage{multirow}%
\usepackage{amsmath,amssymb,amsfonts}%
\usepackage{amsthm}%
\usepackage{mathrsfs}%
\usepackage[title]{appendix}%
\usepackage{xcolor}%
\usepackage{textcomp}%
\usepackage{manyfoot}%
\usepackage{booktabs}%
\usepackage{algorithm}%
\usepackage{algorithmicx}%
\usepackage{algpseudocode}%
\usepackage{listings}%
\usepackage{bbding}
\usepackage{multirow}
\usepackage{array}
\usepackage{makecell}

\begin{document}
	
	\title[Article Title]{Ghost-Stereo: GhostNet-based Cost Volume Enhancement and Aggregation for Stereo Matching Networks}

	\author[1]{\fnm{Xingguang} \sur{Jiang}}\email{jiangxingguang@stu.xhu.edu.cn}
	\author[1]{\fnm{Xiaofeng} \sur{Bian}}\email{bianxiaofeng@stu.xhu.edu.cn}
	\author*[1]{\fnm{Chenggang} \sur{Guo}}\email{chenggang.guo90@hotmail.com}
	\affil[1]{\orgdiv{School of Computer and Software Engineering}, \orgname{Xihua University}, \orgaddress{ \city{Chengdu}, \postcode{610039}, \country{China}}}
	
	\abstract{Depth estimation based on stereo matching is a classic but popular computer vision problem, which has a wide range of real-world applications. Current stereo matching methods generally adopt the deep Siamese neural network architecture, and have achieved impressing performance by constructing feature matching cost volumes and using 3D convolutions for cost aggregation. However, most existing methods suffer from large number of parameters and slow running time due to the sequential use of 3D convolutions. In this paper, we propose Ghost-Stereo, a novel end-to-end stereo matching network. The feature extraction part of the network uses the GhostNet to form a U-shaped structure. The core of Ghost-Stereo is a GhostNet feature-based cost volume enhancement (Ghost-CVE) module and a GhostNet-inspired lightweight cost volume aggregation (Ghost-CVA) module. For the Ghost-CVE part, cost volumes are constructed and fused by the GhostNet-based features to enhance the spatial context awareness. For the Ghost-CVA part, a lightweight 3D convolution bottleneck block based on the GhostNet is proposed to reduce the computational complexity in this module. By combining with the context and geometry fusion module, a classical hourglass-shaped cost volume aggregate structure is constructed. Ghost-Stereo achieves a comparable performance than state-of-the-art real-time methods on several publicly benchmarks, and shows a better generalization ability.}
	
	
	
	\maketitle
	
	\section{Introduction}\label{sec1}
	
    Stereo matching refers to methods that predict a dense disparity map by matching pixels in a pair of stereo-rectified binocular images. According to the formula that depth equals baseline length times focal length divided by disparity, the disparity map can be further converted into a dense depth map. It has wide practical applications in many fields, such as augmented reality, 3D reconstruction, unmanned aerial vehicles and autonomous driving. The current research hotspot of stereo matching methods is to construct an end-to-end dense disparity regression network based on deep neural networks, so as to make full use of the advantages of deep neural networks in automatically learning robust features and highly nonlinear matching functions on large-scale training sets. In the module design of the deep stereo matching network, most methods still use the classic processing flow, namely feature extraction, feature matching, disparity aggregation and optimization.
    
    In recent research, stereo matching methods such as GCNet\cite{kendall2017end}, PSMNet\cite{chang2018pyramid} and GwcNet\cite{guo2019group} have achieved good results. These state-of-the-art networks use Siamese-structured convolutional neural networks (CNNs) to extract deep features and then build a so-called cost volume structure to store matching cost information for each pixel. But this initialized cost volume is not accurate enough. To address this issue, researchers have proposed various cost aggregation methods to aggregate contextual information and penalize noises in the cost volume. For example, GCNet utilized an encoder-decoder module composed of 3D convolutions to regularize a 4D cost volume. Following this idea, PSMNet proposed an hourglass regularization model, and GwcNet further constructed a grouped cost volume based on PSMNet. These methods all stack deep 3D codec structure to aggregate the cost volume. Although achieving good results, 3D convolution takes up a lot of memory and increases the inference time. In order to reduce the memory consumption and calculation cost caused by 3D convolution, GANet\cite{zhang2019ga} designed two guided aggregation layers to replace 3D convolution, and MobileStereoNet \cite{shamsafar2022mobilestereonet} transformed 3D convolution with the idea of depth separability.
    
   	\begin{figure*}
    	\centering
    	\includegraphics[width=1\linewidth]{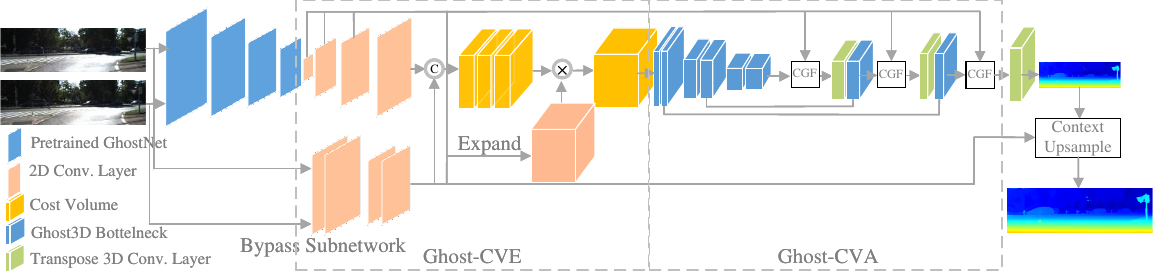}
    	\caption{The proposed Ghost-Stereo stereo matching network structure.}
    	\label{fig1}
    \end{figure*}
    
    In this paper, a lightweight but accurate stereo matching network called Ghost-Stereo is designed, shown in Figure~\ref{fig1}. In this network, the idea of GhostNet\cite{han2020ghostnet} is mainly introduced, which is to generate redundant feature channels for a convolution layer in a linear manner. For the feature extraction part, GhostNet is a lightweight network that reduces the memory consumption and computational cost of the feature extraction process. The UNet\cite{ronneberger2015u} network has powerful multi-scale feature extraction capabilities. By combining the advantages of GhostNet and UNet, a GhostNet U-shaped feature extraction network is constructed. We use the method of GwcNet to construct the cost volume. However, although it contains rich disparity information, it lacks spatial context and multi-scale image information, and continuous use of multiple layers of 3D convolutions will introduce excessive computational cost. Therefore, this paper proposes a cost volume enhancement module based on GhostNet features and a cost volume aggregation module inspired by GhostNet, named Ghost-CVE and Ghost-CVA, respectively. For the Ghost-CVE part, we directly integrate the feature information of the left image into the group-wise cost volume through dimension expansion and broadcast multiplication operations. The goal of Ghost-CVE is to enhance the spatial multi-scale context representation ability of cost volume. For the Ghost-CVA part, a lightweight 3D convolution bottleneck block based on the GhostNet is proposed to reduce the computational complexity in this module. By introducing the context and geometry fusion (CGF) module proposed in CGI-Stereo\cite{xu2023cgi}, we construct a classic hourglass-shaped cost aggregation structure. When trained only on the synthetic SceneFlow\cite{mayer2016large} dataset, our Ghost-Stereo performs well on other real datasets with strong generalization ability. On the KITTI dataset, it is superior to GCNet\cite{kendall2017end}, PSMNet\cite{chang2018pyramid}, GwcNet\cite{guo2019group} and other state-of-the-art methods in terms of accuracy and speed.

    In summary, our main contributions are:
    
	1. A Ghost-CVE module and a Ghost-CVA module are proposed. The disparity representation ability of cost volume can be improved by fusing GhostNet-based multi-scale context and geometric features, thereby reducing the dependence on 3D convolutions and improving the matching accuracy. Ablation experiments demonstrate the effectiveness of this two modules.
	
	2. An accurate and lightweight stereo matching network Ghost-Stereo is designed, which achieves good results on stereo matching benchmarks and shows good generalization ability.
	
	\section{Related Work}\label{sec2}
	Recently, Siamese CNNs have shown great potential for stereo matching tasks. To deal with complex real-world scenes, popular stereo matching methods usually use 2D convolutions to extract robust features to construct cost volume. GCNet\cite{kendall2017end} constructs 4D cost volumes by concatenating left and right CNN features, and then leverages a 3D encoder-decoder network to aggregate and regularize the cost volumes. Following GCNet\cite{kendall2017end}, PSMNet\cite{chang2018pyramid} and GwcNet\cite{guo2019group} utilize a stacked 3D encoder-decoder structure to regularize the cost volume and achieve a large accuracy improvement. However, massively stacked 3D convolutional layers are computationally expensive and memory consuming. To improve the efficiency of cost aggregation, CoEx\cite{bangunharcana2021correlate} proposes a contextual channel aggregation module to strengthen the disparity regularization ability of 3D convolutions. GANet\cite{zhang2019ga} proposes two kinds of guided aggregation layers to replace most 3D convolutions. SMAR-Net\cite{wang2020self} proposes a self-supervised approach to train stereo matching networks. $HD^3$\cite{yin2019hierarchical} proposed a framework suitable for learning probabilistic pixel correspondences to reduce disparity candidates. EdgeStereo\cite{song2020edgestereo} develops a multi-task learning network to predict disparity maps and edge maps, which uses a residual pyramid instead of 3D convolutions. AANet\cite{xu2020aanet} studies sparse points based on intra-scale cost aggregation to alleviate the well-known edge flattening problem under the condition of difference discontinuity. ACVNet\cite{xu2022attention} constructs a new network fused with the main network and achieves competitive accuracy. Raft-Stereo\cite{lipson2021raft} replaces 3D convolution with ConvGRU to regularize disparity. DeepPruner\cite{Duggal2019DeepPruner} utilizes a differentiable module called patch matching to discard most disparity candidates, which learns the range to prune for each pixel. Although these methods have achieved very good accuracy, it is difficult for these methods to achieve a balance between accuracy and running speed at the same time.
	
   	There are also some studies focusing on lightweight stereo matching networks for real-time performance. They usually construct a low-resolution cost volume or a sparse cost volume to greatly reduce computation complexity. For example, StereoNet\cite{khamis2018stereonet}, BGNet\cite{xu2021bilateral}, and ESMNet\cite{guo2019learning} build and aggregate low-resolution cost volumes, then use edge-preserving refinement modules or upsampling modules based on learned bilateral grids to improve prediction accuracy. DecNet\cite{yao2021decomposition} runs dense matching at very low resolutions and uses sparse matching at high resolutions. MobileStereoNet\cite{shamsafar2022mobilestereonet} modifies 3D convolutions using the idea of depthwise separable convolution. Although these methods achieve real-time performance, they lead to a large degree of accuracy loss.
   	
	In the field of lightweight deep neural networks, MobileNet\cite{howard2017mobilenets} adopts depth separable ideas to reduce the amount of parameters, and GhostNet\cite{han2020ghostnet} uses linear methods to generate redundant features, which are widely used in the field of image classification, object detection and other fields. In this paper, we apply the GhostNet idea to the stereo matching network. Different from existing methods, our proposed Ghost-based CVE and CVA modules can adaptively fuse contextual features into the cost volume, which helps to strengthen the accurate estimation of disparity in the cost volume, and replaces the vanilla 3D convolution, reducing memory and computational burden.
	
	\section{Method}\label{sec3}
	The network structure is shown in Figure~\ref{fig1}. Our Ghost-Stereo includes U-shaped multi-scale feature extraction based on GhostNet\cite{han2020ghostnet}, cost volume construction based on GwcNet\cite{guo2019group}, cost volume enhancement module Ghost-CVE and cost aggregation module Ghost-CVA. In this section, we detail the composition of each module sequentially, and explain the loss function at the end.
	
	\subsection{Multi-scale Feature Extraction}\label{subsec1}
	Our multi-scale feature extraction module adopts the U-shaped structure. The encoding part uses a GhostNet\cite{han2020ghostnet} pre-trained on ImageNet-1k\cite{deng2009imagenet} as the backbone network to obtain multi-scale features. The decoding part uses three upsampling convolution blocks consecutively. Each block contains one transposed convolution layer for feature resolution upsampling, and one convolution layer for feature fusion with skipped features from the same level of the encoding part. Let $I_L,I_R\in\mathbb{R}^{H\times W\times3}$ denote a rectified stereo image pair. Multi-scale features are extracted as $f_s\in\mathbb{R}^{\frac{H}{2^s}\times \frac{W}{2^s}\times C_s}$, where $s\in\{2,3,4,5\}$. Besides, we also introduce a shallow bypass sub-network to provide more low-level information. The bypass sub-network consists of two sets of convolutional blocks, each containing two layers of convolutions, which can reduce the feature resolution by half. Finally, we fuse left and right features from the scale $s=2$ and the bypass features to construct the cost volume.
	
	\subsection{Ghost-CVE Module}\label{subsec2}
	We construct the cost volume in the same way as GwcNet\cite{guo2019group}, which means group-wise correlation is computed for each disparity level $d$, i.e.,
	\begin{equation}
	\label{1}
	C_{gwc}(g,d,y,x)=\frac{G}{C_s}\left<f^g_L(y,x),f^g_R(y,x-d) \right>
	\end{equation}
	where $\left<\cdot,\cdot\right>$ represents the inner product of two feature vectors. Left and right fused features are evenly divided into $G$ groups along the feature channel dimension and then normalized by group, denoted as $f^g_L$ and $f^g_R$, respectively. A single 3D convolution layer is applied to preprocess the cost volume.
	
	Although using group-wise correlation preserves rich similarity metric features for 3D aggregation networks, contextual semantic information is still largely lost. We introduce the fused left image features to enhance the spatial context representation ability of cost volume $C_{gwc}$:
 	\begin{equation}
 	\label{2}
 	 C = C_{gwc} \times U(F^{2D}(f_L))
	\end{equation}
	where $\times$ denotes a broadcasted multiplication, $U(\cdot)$ denotes the dimension expand operation, $F^{2D}(\cdot)$ denotes a 2-layer 2D convolution block to transform the dimension of left image feature $f_L$ into $G$ dimension. Finally, we post-process the cost volume $C_{cve}$ using a 3D convolution with a kernel size of $1\times5\times5$.
    
	\subsection{Ghost-CVA Module}\label{subsec3}
	In order to lightweight the 3D convolution based cost aggregation part, we design a new cost volume aggregation module, called Ghost-CVA for short. This module follows the classic hourglass-shaped cost aggregation structure. First, we use three custom 3D convolution aggregation blocks to perform cost regularization on the cost volume, thereby reducing the amount of computation while reducing the resolution of the cost volume. Then, the context and geometry fusion (CGF) module\cite{xu2023cgi}, transposed 3D convolution, and custom 3D convolution aggregation block are used alternately to decode the low-resolution cost volume into a higher resolution scale and introduce multi-scale contextual features into the cost volume. 
	
	\begin{figure}
		\centering
		\includegraphics[width=\linewidth]{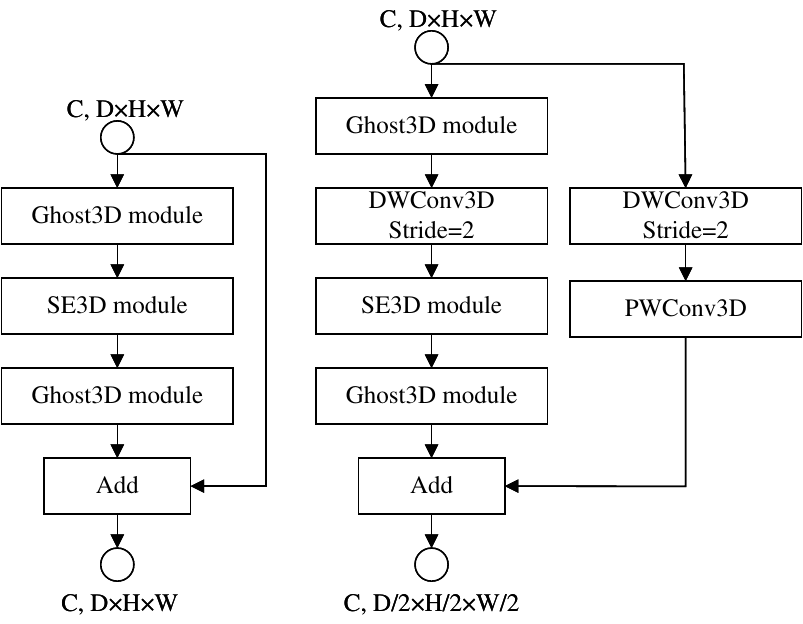}
		\caption{Ghost3D-Bottleneck. Left: Ghost3D-bottleneck with stride=1; right: Ghost3D-bottleneck with stride=2}
		\label{fig2}
	\end{figure}

	\begin{figure}
		\centering
		\includegraphics[width=0.8\linewidth]{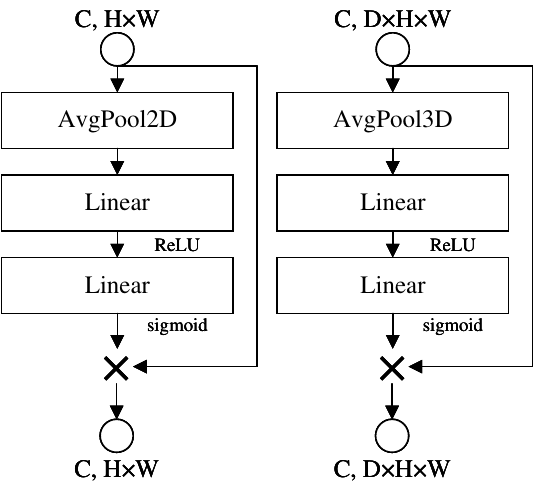}
		\caption{Comparison between SE module (left) and the proposed SE3D module (right).}
		\label{fig3}
	\end{figure}
	
	Specifically, the custom 3D convolution cost aggregation block, called Ghost3D-Bottleneck, is close but different from the original 2D version of Ghost bottleneck, mainly composed of Ghost3D modules, depth-wise 3D convolutions (DWConv3D), point-wise 3D convolutions (PWConv3D) and Squeeze-and-Excitation 3D (SE3D) modules. As shown in Figure~\ref{fig2}, we designed two versions of Ghost3D-Bottleneck, corresponding to the two cases of stride=1 and stride=2 respectively. For the version with stride=1, the first Ghost3D module acts as an extension layer to increase the number of feature channels for the input cost volume. The second Ghost3D module reduces the number of channels to match the shortcut connection. We use the 3D version of the SE module, shown in Figure~\ref{fig3}, to aggregate and enhance the channel features of the cost volume. As for the case where stride=2, the shortcut path is implemented by a downsampling DWConv3D layer and a PWConv3D layer. A depthwise 3D convolution with stride=2 is inserted between the Ghost3D module and the SE3D module.
	
	\begin{figure}
		\centering
		\includegraphics[width=0.8\linewidth]{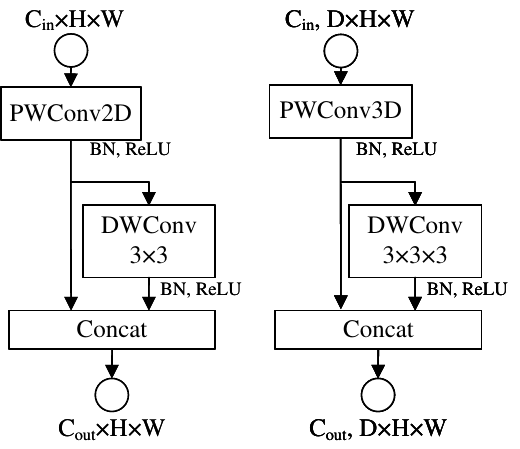}
		\caption{Comparison between Ghost module (left) and the proposed Ghost3D module (right).}
		\label{fig4}
	\end{figure}
	
	As shown in Figure~\ref{fig4}, GhostNet believes that there is a large redundancy in the output feature map of the 2D convolution. Therefore, a Ghost module first uses a PWConv convolution to generate intrinsic feature maps, and then feeds intrinsic feature maps into DWConv to generate more redundant features. The intrinsic features and redundant features are concatenate by channels. Similarly, we believe that this operation can be extended to 3D convolution, that is, the Ghost3D module. In this way, we can reduce the amount of 3D convolution calculations and parameters required by the cost volume aggregation step.
	
	\subsection{Disparity Regression}\label{subsec4}
	We use the top-k softmax disparity regression\cite{bangunharcana2021correlate} to perform disparity estimation on the 4D cost volume after the cost aggregation process. Specifically, only the top-k values at each pixel of the aggregated cost volume are used as weights to alleviate the issue that the regression value diverge from the true value when the matching cost is a multi-modal distribution:
	\begin{equation}
	\label{3}
	\hat{d}=\sum_{i=1}^{k}{ind_i} \times softmax(C(ind_i))
	\end{equation}
	where $ind$ is the set of disparity indices returned by the top-k operation. After regressing the disparity at each pixel location, we get a disparity map of 1/4 resolution of the input image. Similar to the CoEx \cite{bangunharcana2021correlate} method, we reuse the features from the bypass sub-network branch to predict upsampled weights of size $3\times3$ based on the fused left features, and upsample the disparity map resolution to the original image size.

	\subsection{Training Loss}\label{subsec5}
	We train Ghost-Stereo in a supervised end-to-end manner using $smooth_{L1}$ loss function. The loss function is defined as:
	\begin{equation}
	\label{4}
	\mathcal L = \sum_{i=0}^{1}\lambda_ismooth_{L1}(d_{GT,i}-\hat{d}_i)
	\end{equation}
	where $i=0$ refers to a quarter resolution, $i=1$ refers to the full resolution, $d_{GT}$is the ground-truth disparity map.
	
	\section{Experiment}\label{sec5}
	\subsection{Datasets and Metrics}\label{subsec6}
	
	SceneFlow\cite{mayer2016large} is a synthetic dataset containing 35,454 pairs of training images and 4,370 pairs of testing images, which contains dense disparity maps. We use the 'finalpass' version. KITTI 2012\cite{geiger2012we} and KITTI 2015\cite{menze2015object} are datasets for real-world driving scenarios. KITTI 2012 contains 194 pairs of training images and 195 pairs of testing images, and KITTI 2015 contains 200 pairs of training images and 200 pairs of testing images. Both datasets provide sparse ground truth disparity acquired by LIDAR. Middlebury 2014\cite{scharstein2014high} is an indoor dataset that provides 15 pairs of training images and 15 pairs of testing images, where some samples are under inconsistent lighting or color conditions. All images are available in three different resolutions. ETH3D\cite{schops2017multi} is a grayscale dataset containing 27 pairs of training images and 20 pairs of testing images.
	
	We quantitatively evaluate our network with different metrics as follows: End point error (EPE), running time on SceneFlow, the percentage of disparity outliers (D1) for background (bg), foreground (fg) and all pixels (all) and the percentage of pixels with error larger than the specified threshold (3-all) on KITTI, the percentage of pixels with absolute error larger than 2 pixel (bad2.0) on training image pairs from Middlebury 2014, and the 1-pixel error rate (1px) on training image pairs from ETH3D.
	
	\subsection{Experimental details}\label{subsec7}
	
	We implement our network using PyTorch and train the network using NVIDIA RTX3090 gpus. For all experiments, the Adam\cite{kingma2014adam} optimizer with $\beta_1$= 0.9, $\beta_2$= 0.999 is used. The coefficients of the two outputs are set to $\lambda_0$=0.3, $\lambda_1$=1.0. On SceneFlow\cite{mayer2016large}, we first train the Ghost-Stereo network for 20 epochs, and then retrain for 20 epochs with the hyperparameters unchanged. Our initial learning rate is set to 0.001, and the learning rate is half-decayed at epochs of 10, 14, 16, and 18. For KITTI, we train for 600 epochs on weights pre-trained on SceneFlow\cite{mayer2016large}, and we use a mixed training set of KITTI2012\cite{geiger2012we} and KITTI2015\cite{menze2015object}for training. Our initial learning rate is 0.001 and the learning rate decays in half at the 300th epoch.
	
	\begin{table}[]
		\caption{Ablation experiments.}
		\label{ablation}
		\begin{tabular}{c|cc|c}
			\hline
			Method 			& Ghost-CVE 	& Ghost-CVA 	& EPE(px)\\
			\hline
			Baseline    	&   			& 				& 0.96 \\
			+Ghost-CVE      & \Checkmark  	&  				& 0.90 \\
			+Ghost-CVA  	& 			    & \Checkmark 	& 0.76 \\
			Ghost-Stereo    & \Checkmark 	& \Checkmark 	& 0.69 \\
			\hline
		\end{tabular}
	\end{table}
	
	\subsection{Ablation experiments}\label{subsec8}
	In order to verify the effectiveness of the Ghost-CVE and Ghost-CVA modules proposed in this paper, we first perform ablation experiments on the SceneFlow\cite{mayer2016large} test set. The baseline model builds a modified version of GwcNet\cite{guo2019group}. Standard GwcNet\cite{guo2019group} stacks three sets of 3D encoder-decoder structures to regularize the cost volume, but we use only one set of 3D encoder-decoder structure for fair comparison with the proposed network. As shown in Table~\ref{ablation}, our proposed Ghost-CVE and Ghost-CVA can significantly improve the accuracy, and the best performance can be achieved by using these two modules together.

	\subsection{Comparison with State-of-the-art}\label{subsec9}
	\textbf{Scene Flow} As shown in Table~\ref{sceneflow}, our network is compared with a number of state-of-the-art networks. Our network has an EPE of 0.69 and a running time as low as 0.04 seconds.

	\begin{table}[h]
		\caption{Comparison with the state-of-the-art on SceneFlow.}
		\label{sceneflow}
		\begin{tabular}{lcc}
			\hline
			Method 									& EPE(px) 	& Runtime(s) \\
			\hline
			StereoNet\cite{khamis2018stereonet} 	& 1.1   	& \textbf{0.015} \\
			PSMNet\cite{chang2018pyramid} 			& 1.09  	& 0.44\\
			GANet-Deep\cite{zhang2019ga} 			& 0.78  	& 1.8\\
			LEAStereo\cite{cheng2020hierarchical} 	& 0.78  	& 0.3\\
			GwcNet\cite{guo2019group} 				& 0.76  	& 0.36\\
			ESMNet\cite{guo2019learning}            & 0.84  	& 0.06 \\
			DecNet\cite{yao2021decomposition}       & 0.84      & 0.05 \\
			MobileStereoNet3D\cite{shamsafar2022mobilestereonet} & 0.8 & 0.02\\
			Ghost-Stereo(ours) 						& \textbf{0.69}  & 0.04 \\
			\hline
		\end{tabular}
	\end{table}%
	
	\textbf{KTITI} The experimental results are listed in Table~\ref{kitti}. Our method outperforms many lightweight stereo matching methods of the same type, such as ESMNet\cite{guo2019learning}, AANet\cite{xu2020aanet}, CoEx\cite{bangunharcana2021correlate}, MobileStereoNet3D\cite{shamsafar2022mobilestereonet}. These methods all build cost volumes and use 3D convolutions for cost volume regularization, but our method has higher accuracy and faster running time than theirs. Methods with higher accuracy than our network have longer inference times. For example, our network is nearly 50 times faster than GANet\cite{zhang2019ga}. Methods with slightly faster run time than our method have slightly worse accuracy than our method, such as CoEx \cite{bangunharcana2021correlate} and MobileStereoNet3D\cite{shamsafar2022mobilestereonet}. It can be seen that our network achieves a balance between accuracy and speed.
	
	\begin{table*}[h]
		\caption{Comparison with the state-of-the-art on KITTI datasets.}
		\label{kitti}
		\begin{tabular}{clcccccc}
			\hline
			Target & Method & \multicolumn{2}{c}{KITTI2012} & \multicolumn{3}{c}{KITTI2015} & Runtime(s) \\
			       							   &           & 3-noc & 3-all& D1-bg& D1-fg& D1-all & \\
			\hline
			\multirow{6}{*}{\rotatebox{90}{\textit{Accuracy}}}
			& GCNet\cite{kendall2017end}		& 1.77  & 2.30 & 2.02 & 5.58 & 2.61  & 0.9 \\
			& SegStereo\cite{yang2018segstereo} & 1.68 	& 2.03 & 1.88 & 4.07 & 2.25  & 0.6\\
			& GwcNet\cite{guo2019group} 		& 1.52  & 1.98 & 1.74 & 3.93 & 2.11  & 0.32\\
  			& PSMNet\cite{chang2018pyramid} 	& 1.49 	& 1.89 & 1.86 & 4.62 & 2.32  & 0.41\\
			& SSPCVNet\cite{wu2019semantic} 	& 1.47 	& 1.90 & 1.75 & 3.89 & 2.11  & 0.9\\
  			& GANet\cite{zhang2019ga} 			& 1.19 	& 1.60 & 1.48 & 3.46 & 1.81  & 1.8\\
			\hline
			\multirow{8}{*}{\rotatebox{90}{\textit{Speed}}}
			& ESMNet\cite{guo2019learning} 			& 2.08 	& 2.53 & 2.57 & 4.86 & 2.95  & 0.067 \\
			& DeepPrunerFast\cite{Duggal2019DeepPruner} 	& - 	& -    & 2.32 & 3.91 & 2.56  & 0.062\\
		   	& AANet\cite{xu2020aanet}				& 1.91 	& 2.42 & 1.99 & 5.39 & 2.55  & 0.062 \\
  			& DecNet\cite{yao2021decomposition} 	& - 	& -    & 2.07 & 3.87 & 2.37  & 0.05\\
  			& BGNet\cite{xu2021bilateral} 		& - 	& -    & 1.81 & 4.09 & 2.19 & 0.032\\
  			& CoEx\cite{bangunharcana2021correlate} & 1.55 & 1.93 & 1.79 & 3.82 & 2.13 & 0.027 \\
  			& MobileStereoNet3D\cite{shamsafar2022mobilestereonet} & - & - & 1.75 & 3.87 & 2.10 & 0.017\\
  			& Ghost-Stereo(ours) 					& 1.45 & 1.80 & 1.71 & 3.77 & 2.05 & 0.037\\
			\hline
		\end{tabular}
	\end{table*}

\begin{figure}
		\centering
		\includegraphics[width=1\linewidth]{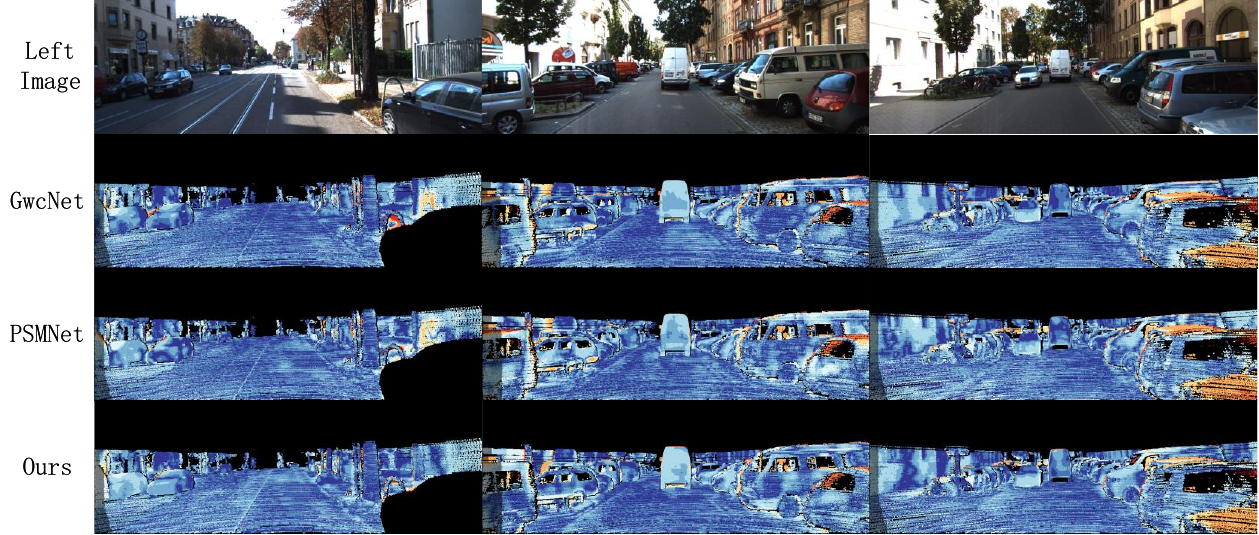}
		\caption{Visualization of disparity prediction error map on the KITTI 2015 test set.}
		\label{kitti15}
	\end{figure}

	Figure~\ref{kitti15} further visualizes the disparity prediction error map on the KITTI 2015 test set. Our method outperforms the compared baseline models GwcNet\cite{guo2019group} and PSMNet\cite{chang2018pyramid} based on 3D convolutions and hourglass aggregation structures, especially with lowers errors near object contours.
	
	\subsection{Generalization performance}\label{subsec11}
\begin{figure}
		\centering
		\includegraphics[width=1\linewidth]{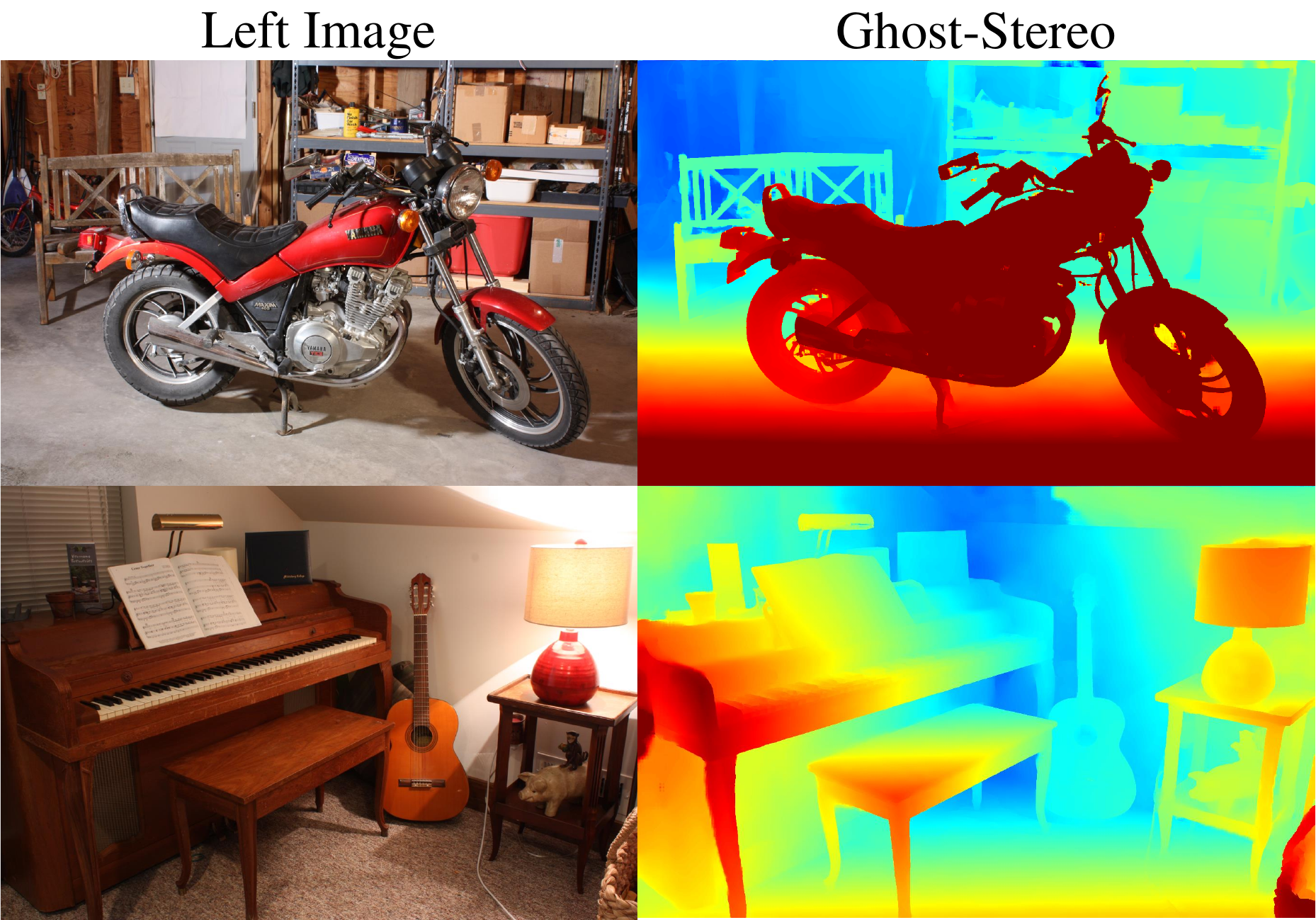}
		\caption{Visualization of the generalization results on Middlebury 2014.}
		\label{genfig}
	\end{figure}

	After our network is pretrained on the synthetic dataset SceneFlow, the model generalization performance is evaluated directly on the training set of Middlebury 2014 and ETH3D. From Table~\ref{gentable}, we can see that our network has better generalization performance than most networks on these two real-world datasets, and has strong generalization performance. In Middlebury 2014, our generalization performance metric is the best. The comparative results of our model on the ETH3D dataset get the fourth place. We give two generalization visualization results of our network in Figure~\ref{genfig}.

	\begin{table}
		\caption{Generalization experiments.}
		\begin{tabular}{lcc}
			\hline
			Method & \makecell[c]{Middlebury 2014 \\bad2.0} & \makecell[c]{ETH3D\\1px} \\
			\hline
			DeepPrunerFast\cite{Duggal2019DeepPruner}	& 38.7  & 36.8 \\
			BGNet\cite{xu2021bilateral} 		  	& 24.7  & 22.6 \\
			GANet\cite{zhang2019ga} 				& 20.3  & 14.1 \\
			PSMNet\cite{chang2018pyramid} 			& 15.8  & 9.8 \\
			STTR\cite{li2021revisiting} 				& 15.5  & 17.2 \\
			CFNet\cite{shen2021cfnet} 				& 15.4  & \textbf{5.3} \\
			CoEx\cite{bangunharcana2021correlate} 	& 14.5  & 9 \\
			DSMNet\cite{zhang2020domain} 		& 13.8  & 6.2\\
			CGI-Stereo\cite{xu2023cgi} 				& 13.5  & 6.3 \\
			Ghost-Stereo(ours) 						& \textbf{11.4}  & 7.3 \\
			\hline
		\end{tabular}
		\label{gentable}
		\end{table}
	
	\section{Conclusion}\label{sec6}
	We propose Ghost-Stereo, a stereo matching network architecture based on the idea of GhostNet to reduce the redundancy of 3D convolutions, which achieves real-time performance and is more accurate than existing lightweight stereo matching methods. By redesigning the conventional hourglass-type cost aggregation structure based on 3D convolution, the Ghost-CVE module and Ghost-CVA module proposed by us can adaptively integrate the context and geometric information into the cost volume, which improves the running speed and reduces the number of parameters. A balance between accuracy and runtime is achieved. Future work will further consider improving the generalization performance of the network in real scenarios.

	\section*{Declarations}
	\textbf{Funding} This work was supported by the talent introduction project of Xihua University under grant no.Z212028.
	
	\textbf{Conflict of interest} The authors have no competing interests to declare that are relevant to the content of this article.
	
	\textbf{Code availability} Code will be available on request.
	
	\textbf{Authors' contributions} X. Jiang: Conceptualization, Methodology, Software and experiments, Writing - original draft preparation and editing. X. Bian: Methodology, Software and experiments. C. Guo: Conceptualization, Writing - review and editing.

	\bibliography{ghost-stereo}

\end{document}